# Towards Intelligent Energy Security: A Unified Spatio-Temporal and Graph Learning Framework for Scalable Electricity Theft Detection in Smart Grids

[1]AbdulQoyum A. OLOWOOKERE*, [2]Usman A. OGUNTOLA, [3]Ebenezer. Leke ODEKANLE, [4]Maridiyah A. MADEHIN, [5]Aisha A. ADESOPE

[1,3,4]Department of Chemical and Petroleum Engineering, Abiola Ajimobi Technical University, Ibadan, Oyo State, Nigeria.

[2,5]Department of Computer Science, Abiola Ajimobi Technical University, Ibadan, Oyo State, Nigeria

[1]abdulqoyumadegoke@gmail.com , [2]uoguntola@gmail.com, [3]odekanleebenezer@gmail.com, [4]madehinadebukolamordiyah@gmail.com,[5]adesopeaisha52@gmail.com.

**Abstract**

Electricity theft and non-technical losses (NTLs) remain critical challenges in modern smart grids, causing significant economic losses and compromising grid reliability. This study introduces the SmartGuard Energy Intelligence System (**SGEIS**), an integrated artificial intelligence framework for electricity theft detection and intelligent energy monitoring. The proposed system combines supervised machine learning, deep learning-based time-series modeling, Non-Intrusive Load Monitoring (NILM), and graph-based learning to capture both temporal and spatial consumption patterns.

A comprehensive data processing pipeline is developed, incorporating feature engineering, multi-scale temporal analysis, and rule-based anomaly labeling. Deep learning models, including Long Short-Term Memory (LSTM), Temporal Convolutional Networks (TCN), and Autoencoders, are employed to detect abnormal usage patterns. In parallel, ensemble learning methods such as Random Forest, Gradient Boosting, XGBoost, and LightGBM are utilized for classification. To model grid topology and spatial dependencies, Graph Neural Networks (GNNs) are applied to identify correlated anomalies across interconnected nodes. The NILM module enhances interpretability by disaggregating appliance-level consumption from aggregate signals.

Experimental results demonstrate strong performance, with Gradient Boosting achieving a ROC-AUC of 0.894, while graph-based models attain over 96% accuracy in identifying high-risk nodes. The hybrid framework improves detection robustness by integrating temporal, statistical, and spatial intelligence. Overall, SGEIS provides a scalable and practical solution for electricity theft detection, offering high accuracy, improved interpretability, and strong potential for real-world smart grid deployment.

**1.0 Introduction**

The modernization of electrical power systems into smart grids has significantly enhanced energy management through advanced monitoring, automation, and data-driven decision-making. Despite these advancements, electricity theft and non-technical losses remain critical challenges affecting the efficiency, reliability, and economic sustainability of power systems. Studies have shown that electricity theft contributes substantially to energy losses, particularly in developing regions, thereby posing serious operational and financial concerns for utility providers [1], [2].

Traditional electricity theft detection approaches, including manual inspection and rule-based monitoring systems, are often inefficient, labor-intensive, and incapable of handling the complexity and scale of modern smart grid data. As a result, there has been a paradigm shift toward the use of machine learning (ML) techniques for automated and scalable detection of fraudulent consumption patterns. Several studies have demonstrated the effectiveness of ML models such as Random Forest, Gradient Boosting, and ensemble learning techniques in identifying anomalous electricity usage behaviors [5], [15], [16].

In recent years, deep learning (DL) approaches have further improved detection performance by capturing complex nonlinear and temporal relationships in electricity consumption data. Models such as Long Short-Term Memory (LSTM), bidirectional recurrent neural networks (BiGRU–BiLSTM), and convolutional neural networks (CNNs) have been successfully applied for electricity theft detection and anomaly identification in smart grids [4], [11], [24]. These models leverage sequential patterns in high-resolution time-series data to enhance predictive accuracy and robustness.

Furthermore, hybrid and ensemble-based approaches have been widely adopted to improve detection performance, particularly in handling imbalanced datasets and noisy real-world data. Techniques combining multiple classifiers and optimization strategies, such as AdaBoost and genetic algorithms, have demonstrated significant improvements in classification accuracy and detection reliability [7], [14], [20].

Despite these advancements, most existing studies primarily focus on temporal analysis while neglecting the spatial and structural characteristics of power distribution networks. In real-world smart grids, electricity flows through interconnected infrastructures consisting of transformers, feeders, and consumer nodes, where anomalies may propagate across the network. To address this limitation, recent research has explored graph-based learning techniques, including Graph Neural Networks (GNNs), to model spatial dependencies and detect coordinated theft patterns within grid topologies [22].

Additionally, the integration of renewable energy sources and environmental variables introduces further complexity into electricity consumption patterns. Factors such as solar and wind generation, temperature, and humidity significantly influence load dynamics, thereby necessitating more comprehensive and adaptive detection frameworks [23]. Moreover, emerging technologies such as federated learning and blockchain-based systems have been proposed to enhance privacy, security, and scalability in smart grid applications [10], [12], [21].

In response to these challenges, this study proposes the SmartGuard Energy Intelligence System (SGEIS), an advanced AI-driven framework for real-time grid monitoring, anomaly detection, and electricity theft identification. The proposed system integrates multi-source smart grid data, including electrical parameters, renewable energy inputs, and environmental factors, to provide a holistic view of grid operations.

The SGEIS framework combines multi-scale time-series analysis, hybrid anomaly detection models (LSTM, Temporal Convolutional Networks, and Autoencoders), and advanced machine learning classifiers for robust detection of anomalous consumption patterns. Furthermore, graph-based learning techniques are incorporated to capture spatial dependencies and identify propagation patterns within transformer–meter networks.

The main contributions of this study are summarized as follows:

  i. Development of a comprehensive smart grid intelligence framework integrating electrical, environmental, and renewable energy data.
 ii. Implementation of a hybrid anomaly detection approach combining LSTM, TCN, and Autoencoder models.
iii. Design of a rule-based and machine learning-driven labeling mechanism for electricity theft detection.
 iv. Integration of ensemble-based supervised learning models for improved classification performance.
  v. Application of graph-based learning techniques to capture spatial dependencies and detect network-level anomalies.
 vi. Extensive experimental evaluation demonstrating the effectiveness and scalability of the proposed framework.

The remainder of this paper is organized as follows: Section 2 reviews related literature on electricity theft detection and smart grid analytics. Section 3 presents the proposed methodology and system architecture. Section 4 discusses experimental results and performance evaluation. Finally, Section 5 concludes the study and outlines future research directions.

**2.0 Related Studies**

Recent studies on electricity theft detection have increasingly focused on the application of artificial intelligence, machine learning, deep learning, and privacy-preserving smart grid frameworks to improve the identification of anomalous and fraudulent energy consumption patterns. Earlier approaches largely relied on conventional machine learning models such as decision trees, support vector machines, random forests, and boosting algorithms to classify normal and suspicious electricity usage behaviors. These methods demonstrated promising detection capability, especially when combined with feature engineering and data resampling strategies to handle class imbalance and noisy meter readings [1], [2], [5], [15], [20].

As research advanced, deep learning-based methods gained greater attention because of their ability to automatically learn complex nonlinear and temporal dependencies from smart meter data. Recurrent architectures such as LSTM, BiGRU, and BiLSTM, as well as convolutional and transformer-based models, have been employed to capture sequential consumption patterns and detect subtle anomalies associated with electricity theft [4], [11], [22], [24]. These approaches generally reported improved predictive performance over conventional machine learning models, particularly in large-scale smart grid environments.

In addition, recent studies have explored ensemble and hybrid methods that combine multiple algorithms to improve robustness, generalization, and detection accuracy. Optimization-driven frameworks, such as genetic algorithm-enhanced boosting methods, have also been introduced to strengthen anomaly detection performance in complex electricity consumption settings [7], [14], [16]. At the same time, smart grid research has expanded beyond pure theft detection to include privacy-preserving federated learning, blockchain-supported resilience models, and IoT-enabled surveillance systems, all of which aim to strengthen security, scalability, and trustworthiness in modern power systems [10], [12], [19], [21].

Despite these contributions, important limitations remain in the literature. Many existing studies are restricted to either purely temporal consumption analysis or flat tabular classification, with limited consideration of grid topology, renewable energy integration, and environmental influences on energy demand. Furthermore, several methods focus on a single detection model rather than a unified framework capable of combining anomaly detection, supervised classification, and graph-based intelligence. These gaps motivate the development of the proposed SmartGuard Energy Intelligence System (SGEIS), which integrates multi-scale temporal modeling, hybrid machine learning, and graph-based analysis for more comprehensive electricity theft detection and grid intelligence.

**Table 1. Summary of Related Studies on Electricity Theft Detection and Smart Grid Intelligence**

| Author(s) | Study Purpose & Application Area | Methodology | Dataset | Technique | Key Findings | Limitations |
|---|---|---|---|---|---|---|
| Kabir et al. (208) | Electricity theft detection in smart grids | Ensemble ML + Explainable AI | Smart meter data | Ensemble (RF, XGBoost) + XAI | High detection accuracy with interpretability | Limited temporal modeling |
| Iftikhar et al. (2024) | ML-based electricity theft detection | Supervised learning framework | Smart grid dataset | Random Forest, SVM | Effective classification of abnormal usage patterns | Ignores spatial dependencies |
| Munawar et al. (2022) | Deep learning-based theft detection | Hybrid DL architecture | Time-series smart meter data | BiGRU–BiLSTM | Strong temporal pattern learning | High computational cost |
| Kawoosa et al. (2023) | Ensemble learning for theft detection | ML ensemble approach | Smart meter data | Ensemble methods | Improved classification performance | Limited time-series modeling |

| Author(s) | Study Purpose & Application Area | Methodology | Dataset | Technique | Key Findings | Limitations |
|---|---|---|---|---|---|---|
| Qu et al. (2021) | Energy anomaly detection | Optimization-based ML | Building energy dataset | GA + AdaBoost | Enhanced anomaly detection accuracy | Not tailored to smart grid topology |
| Wen et al. (2022) | Privacy-preserving theft detection | Federated learning framework | Distributed smart grid data | Federated learning | Ensures data privacy in detection systems | Communication overhead |
| Oguntola et al. (2026) | Spatio-temporal electricity theft detection | Hybrid graph + temporal modeling | Smart grid network data | GNN + LSTM | Captures spatial and temporal dependencies effectively | High computational complexity |

## 3.0 Methodology

This study presents the SmartGuard Energy Intelligence System (SGEIS), a unified artificial intelligence framework designed for real-time electricity theft detection, anomaly identification, and smart grid intelligence. The proposed methodology integrates multi-source data processing, advanced feature engineering, hybrid machine learning models, and graph-based learning techniques to capture both temporal and spatial dynamics within the power grid.

## 3.1 System Overview

The SGEIS framework follows a multi-stage pipeline consisting of data acquisition, preprocessing, feature engineering, anomaly detection, classification, and graph-based analysis. The system utilizes high-resolution smart grid data, including electrical parameters (voltage, current, power consumption), renewable energy sources (solar and wind), environmental variables (temperature and humidity), and operational indicators (fault signals and overload conditions).

The overall architecture is designed to ensure scalability, robustness, and real-time applicability, enabling efficient monitoring of grid behavior and detection of abnormal consumption patterns.

### 3.2 Data Preprocessing and Temporal Structuring

The dataset is first subjected to cleaning and validation processes to ensure data integrity. Missing values are handled using domain-aware imputation techniques, where numerical features are filled using median values and categorical features using mode values. The timestamp attribute is converted into a datetime format and used as the index to enable time-series analysis.

To ensure uniform temporal resolution, the dataset is resampled at 15-minute intervals using mean aggregation. This allows consistent representation of consumption patterns and supports downstream time-series modeling.

### 3.3 Feature Engineering

To enhance model performance and capture complex grid dynamics, multiple features are engineered based on electrical, temporal, and environmental relationships.

#### 3.3.1 Supply–Demand Modeling

To capture the relationship between energy supply and consumption within the grid, several derived features were computed.

The total supply is defining using equation 1:

$$Total\ Supply = Grid\ Supply + Solar\ Power + Wind\ Power \quad (1)$$

where Grid Supply represents electricity from the main distribution network, while Solar Power and Wind Power represent renewable energy contributions.

The total demand is represented by the measured power consumption shown in equation 2:

$$Total\ Demand = Power\ Consumption \quad (2)$$

To quantify discrepancies between supply and demand, the imbalance is calculated using equation 3:

$$Imbalance = Total\ Supply - Total\ Demand \qquad (3)$$

This feature is particularly important because significant deviations between supply and demand may indicate abnormal conditions such as energy leakage, metering errors, or electricity theft.

In addition, the loss percentage is computed using equation 4:

$$Loss\ Percentage = \frac{Grid\ Supply - Power\ Consumption}{Grid\ Supply} \qquad (4)$$

Multi-Scale Temporal Features

To effectively capture both short-term fluctuations and long-term consumption trends, multi-scale temporal features were generated using rolling window statistics. This approach enables the model to analyze consumption behavior at different time horizons, which is essential for detecting both sudden anomalies and gradual deviations.

Three temporal windows were considered: Short-term window (1 hour), Medium-term window (6 hours), and Long-term window (24 hours).

For each window, statistical measures such as the mean and standard deviation were computed using equation 5 and 6 respectively.

$$\mu = \frac{1}{n}\sum_{i=1}^{n} x_i \qquad (5)$$

$$\sigma = \sqrt{\frac{1}{n}\sum_{i=1}^{n}(x_i - \mu)^2} \qquad (6)$$

These features enable the model to detect sudden deviations and gradual behavioral changes.

### 3.3.3 Environmental and Derived Features

Additional features are constructed by combining consumption with environmental variables:

Temperature-adjusted consumption

Humidity-adjusted consumption

Price-weighted consumption

Apparent power is also computed using equation 7:

$$S = \sqrt{P^2 + Q^2} \qquad (7)$$

## 3.4 Rule-Based Labeling Mechanism

A hybrid labeling strategy is implemented to generate ground truth labels for electricity theft detection. Data points are classified as anomalous (Label = 1) if any of the following conditions are met:

  i. Transformer–meter imbalance exceeds threshold
  ii. Voltage fluctuation exceeds acceptable range
  iii. Power factor falls below 0.85
  iv. Fault inconsistencies are detected

Otherwise, the data is labeled as normal (Label = 0). This approach simulates real-world detection rules used in utility systems.

## 3.5 Time-Series Anomaly Detection

To detect anomalies in electricity consumption patterns, deep learning-based time-series models were employed. These models are particularly effective in capturing temporal dependencies and identifying irregular patterns that may indicate electricity theft. In this study, three complementary models were utilized: Long Short-Term Memory (LSTM), Temporal Convolutional Network (TCN), and Autoencoder.

### 3.5.1 Long Short-Term Memory (LSTM)

The LSTM model is designed to capture long-term dependencies in sequential data by maintaining a memory of previous states. The hidden state of the LSTM at time ( t ) is given by equation 8:

$$h_t = \sigma(W_h h_{t-1} + W_x x_t + b) \tag{8}$$

where $x_t$ represents the input at time $t$, $h_{t-1}$ is the previous hidden state, $W_h$ and $W_x$ are weight matrices, $b$ is a bias term, and $\sigma$ is a nonlinear activation function.

The LSTM model learns temporal patterns in electricity consumption and predicts expected behavior. Anomalies are identified by computing the deviation between predicted and actual values. Large deviations indicate abnormal consumption patterns, which may be associated with electricity theft.

### 3.5.2 Temporal Convolutional Network (TCN)

The Temporal Convolutional Network (TCN) is a convolution-based model that captures long-range temporal dependencies using dilated causal convolutions. The output at time $t$ is defined as:

$$y(t) = \sum_{k=0}^{K} w_k \cdot x(t - d \cdot k) \tag{11}$$

where (d) is the dilation factor.

Unlike recurrent models, TCN processes sequences in parallel and can efficiently capture long-term dependencies. This makes it particularly suitable for detecting subtle and delayed anomalies in electricity consumption patterns.

### 3.5.3 Autoencoder

The Autoencoder is an unsupervised neural network that learns to reconstruct input data. It consists of an encoder that compresses the input and a decoder that reconstructs it. The anomaly score is computed using the reconstruction error:

$$Reconstruction\ Error = \|X - \hat{X}\| \tag{12}$$

where $X$ is the original input and $\hat{X}$ is the reconstructed output.

High reconstruction error indicates that the input pattern deviates from learned normal behavior, thereby signaling a potential anomaly. This makes the Autoencoder effective for detecting previously unseen or unknown anomaly patterns.

### 3.5.4 Ensemble Anomaly Detection

The outputs of LSTM, TCN, and Autoencoder are combined using a voting mechanism. A data point is flagged as anomalous if at least two models agree.

**Supervised Classification Models**

To classify electricity theft, multiple machine learning models are trained and evaluated: Random Forest, Gradient Boosting, XGBoost, and LightGBM

The models are trained using engineered features, and performance is evaluated using accuracy, F1-score, and ROC-AUC metrics.

**Graph-Based Learning (GNN)**

To capture spatial dependencies within the grid, a graph-based representation is constructed where: Nodes represent meters and transformers, and Edges represent electrical connections.

Graph Neural Networks (GNNs) are used to propagate information across nodes:

$$H^{(l+1)} = \sigma(AH^{(l)}W^{(l)}) \quad (13)$$

where $A$ is the adjacency matrix and $H$ represents node features.

Models such as Graph Convolutional Networks (GCN) and Graph Attention Networks (GAT) are implemented to detect coordinated anomalies across the network.

### 3.8 Hybrid Detection Framework

The final detection system integrates:

i. Time-series anomaly scores
ii. Supervised classification probabilities
iii. Graph-based node risk scores

These components are combined to generate a unified risk score for each node, enabling robust detection of electricity theft and the propagation of anomalies.

**Evaluation Metrics**

The performance of the proposed system is evaluated using: Accuracy, Precision, Recall, F1-score, and ROC-AUC

These metrics provide a comprehensive assessment of model effectiveness, particularly in imbalanced datasets typical of electricity theft detection.

**Implementation Environment**

The system is implemented using Python with libraries including: Pandas and NumPy for data processing, Scikit-learn for machine learning models, TensorFlow/Keras for deep learning, and PyTorch and PyTorch Geometric for graph modeling

The framework is designed to support scalable deployment in real-world smart grid environments.

### 4.0 Results and Discussion
### 4.1 Introduction

This section presents a comprehensive evaluation of the SmartGuard Energy Intelligence System (SGEIS), focusing on classification performance, anomaly detection effectiveness, and graph-based intelligence. Beyond reporting results, this section provides an in-depth interpretation of model behavior and its implications for real-world smart grid applications.

**4.2 Dataset Characteristics and Label Distribution**

As shown in Fig. 1, the processed dataset is clearly imbalanced, with the normal class (Label 0) accounting for approximately 81.75% of the observations, while the theft/anomaly class (Label 1) represents only about 18.8%. This means that the number of normal consumption records is substantially higher than the number of suspicious or anomalous records. Such a distribution is expected in practical smart grid environments because most consumers operate under legitimate conditions, whereas electricity theft and related anomalous events occur less frequently. Therefore, the imbalance observed in this study is not an artificial issue but rather a realistic reflection of field conditions in electricity monitoring systems.

From a modeling perspective, however, this imbalance creates an important challenge. Since classification algorithms are trained to minimize overall prediction error, they naturally become biased toward the majority class. In this case, a model may correctly classify a large proportion of normal samples and still achieve a high overall accuracy, even if it fails to detect many theft cases. For example, if a classifier predicts most observations as normal, its accuracy may remain high simply because normal cases dominate the dataset. This can create a misleading impression of good model performance, while in reality the model may have weak sensitivity to the minority theft class, which is the class of greatest operational importance.

For this reason, accuracy alone is insufficient for evaluating electricity theft detection systems under imbalanced conditions. More informative metrics such as precision, recall, F1-score, and ROC-AUC are required. Precision indicates how many of the samples predicted as theft are truly theft cases, while recall measures how many actual theft cases are successfully detected by the model. The F1-score provides a balanced summary of these two measures, making it particularly useful when the minority class is critical. Similarly, ROC-AUC evaluates the model's ability to distinguish between normal and anomalous classes across different classification thresholds, thereby providing a more robust measure of separability than accuracy alone. In the context of smart grid security, these metrics are especially important because the primary objective is not merely to classify the majority of samples correctly, but to reliably identify the relatively small number of fraudulent cases without generating excessive false alarms.

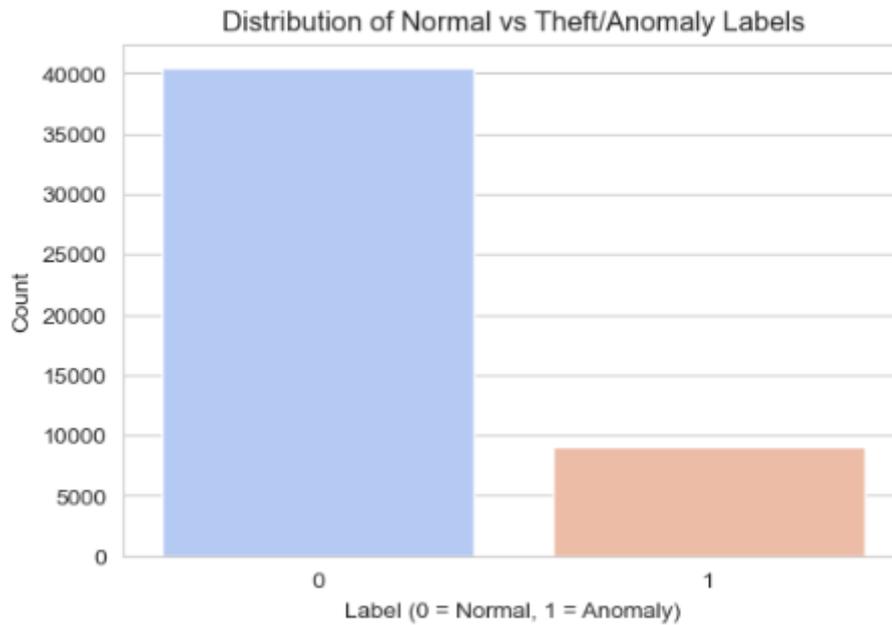

**Fig. 1.** Distribution of normal and anomalous samples.

### 4.3 Exploratory Data Analysis (EDA)

The distribution of power consumption values is presented in Fig. 2. It can be observed that anomalous samples significantly overlap with normal consumption values, indicating that electricity theft does not necessarily manifest as extreme or easily separable values. Instead, theft-related behavior often appears as subtle deviations within normal operating ranges. This makes simple threshold-based detection ineffective and highlights the need for advanced machine learning models capable of capturing complex and nonlinear patterns in the data.

The temporal behavior of the system is further illustrated in Fig. 3. The time-series plot reveals irregular fluctuations, including abrupt spikes and sudden drops in consumption. Such patterns are commonly associated with tampering activities, including meter bypassing and illegal connections. These findings emphasize the importance of temporal models such as Long Short-Term Memory (LSTM) and Temporal Convolutional Networks (TCN), which are specifically designed to capture sequential dependencies and detect abnormal variations over time.

The interrelationship between features is shown in Fig. 4. The correlation heatmap indicates strong dependencies among key variables such as voltage, current, and power consumption. This confirms that electricity usage is influenced by multiple interacting factors rather than a single variable. Consequently, the inclusion of engineered features, such as apparent power and supply–demand imbalance, provides a more informative representation of system behavior and improves model performance.

Further comparison between normal and anomalous samples is presented in Fig. 5. Although anomalous cases exhibit greater variability, there remains a noticeable overlap between the two classes. This observation reinforces the complexity of electricity theft detection, as fraudulent behavior may closely resemble legitimate consumption patterns. Therefore, accurate detection requires sophisticated modeling approaches that can distinguish subtle differences within overlapping distributions.

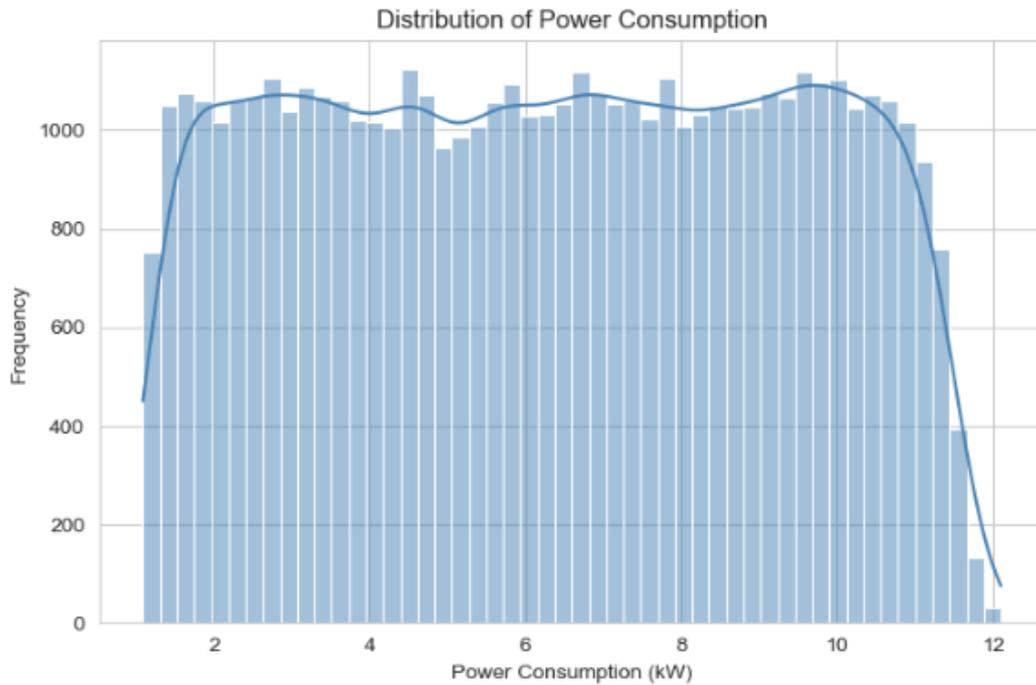

**Fig. 2.** Distribution of power consumption.

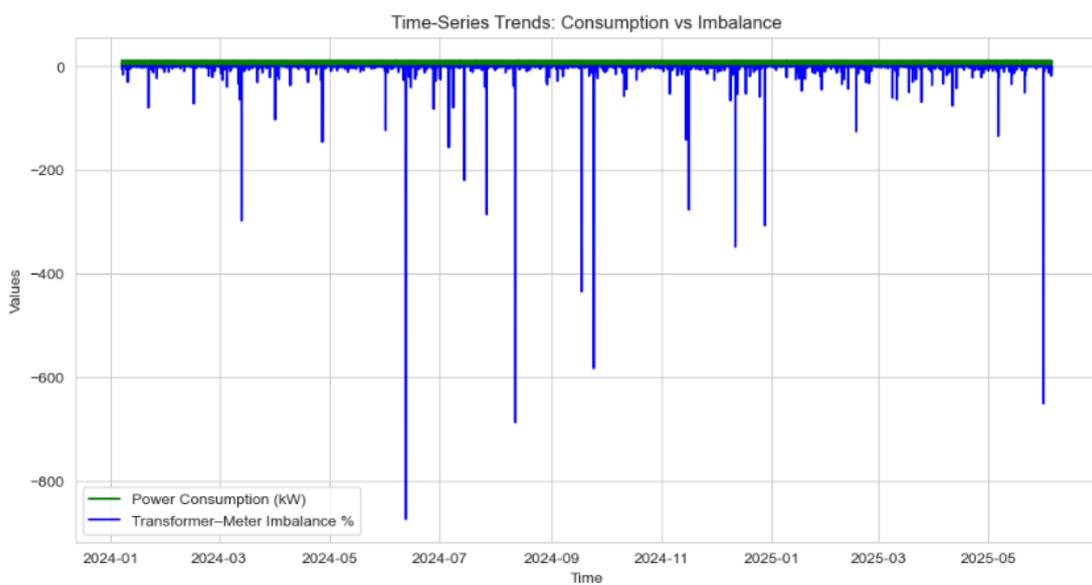

**Fig. 3.** Time-series consumption and imbalance patterns.

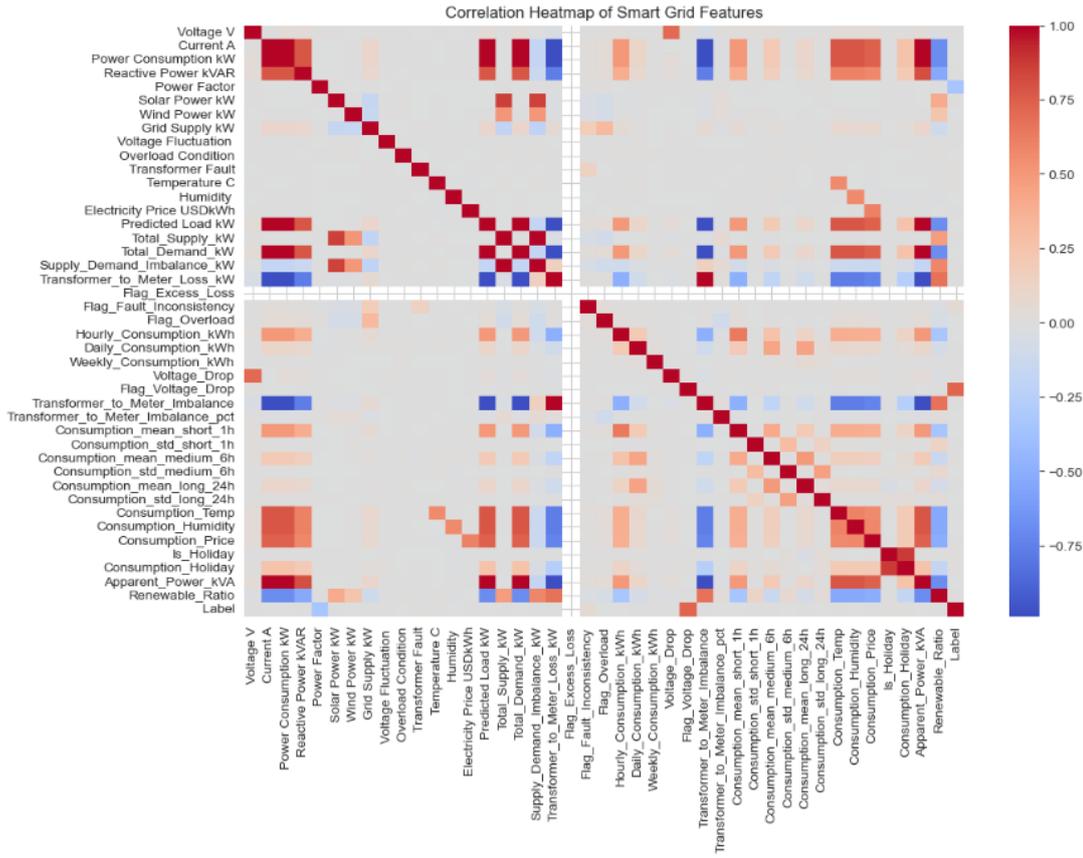

**Fig. 4.** Correlation heatmap.

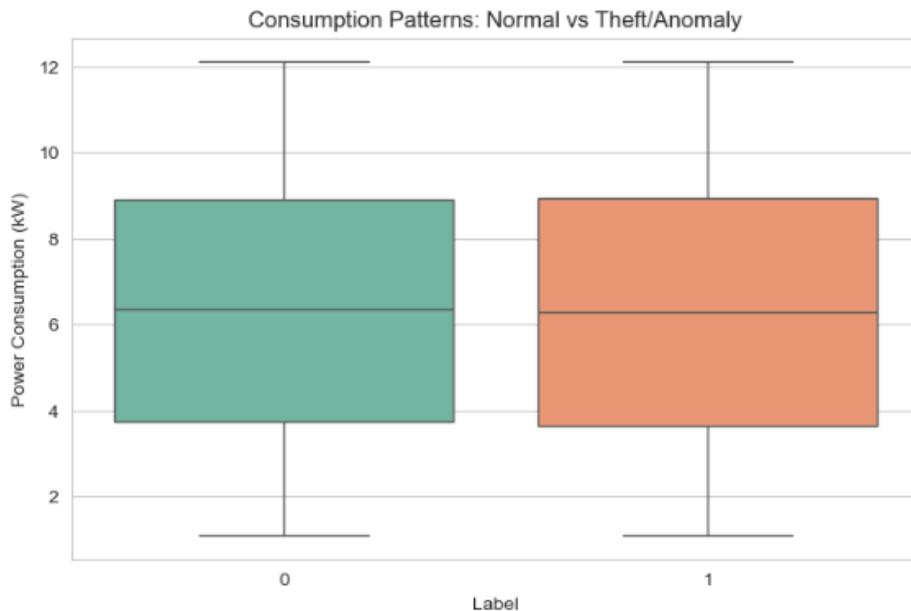

**Fig. 5.** Normal vs anomaly comparison of Power Consumption.

## 4.4 Performance of Supervised Machine Learning Models

The performance of the supervised learning models is summarized in Table 2. The results show that all models achieve comparable accuracy (≈0.91), indicating strong general classification performance. However, a deeper analysis reveals that Gradient Boosting achieves the highest ROC-AUC, suggesting superior capability in distinguishing between normal and anomalous classes.

The similarity in performance across models indicates that the engineered features are highly informative. However, the relatively moderate F1-scores suggest that detecting the minority class (electricity theft) remains challenging. This is primarily due to overlapping feature distributions and class imbalance.

Non-Intrusive Load Monitoring (NILM) was incorporated into the SGEIS framework to identify appliance-level energy usage patterns from aggregated consumption signals. This component is particularly important because abnormal appliance signatures can serve as indirect indicators of unauthorized usage behavior, bypass activities, or hidden load manipulation. In this study, appliance disaggregation was demonstrated using the refrigerator load profile, since refrigerators typically exhibit relatively stable cyclic operating behavior and are therefore suitable for evaluating pattern separation performance.

As shown in Fig. 7, the refrigerator load profile exhibits periodic activation and deactivation cycles consistent with compressor-based operation. The disaggregation output indicates that the proposed framework is capable of separating appliance-level signatures from aggregated energy signals with reasonable temporal consistency. This is important because, under normal operating conditions, appliance behavior follows repetitive and physically interpretable patterns. Any substantial deviation from these regular cycles may indicate appliance misuse, abnormal operating conditions, or the presence of hidden loads connected through unauthorized means.

The refrigerator disaggregation result also demonstrates the ability of the framework to move beyond bulk consumption analysis toward device-level intelligence. This strengthens the practical value of the proposed system because utilities and grid operators are often interested not only in whether abnormal activity occurred, but also in which category of load contributed to the irregularity. In the context of electricity theft detection, NILM therefore provides an additional interpretability layer by linking abnormal aggregate behavior to identifiable appliance signatures.

**Table 2. Performance comparison of supervised models**

| Model | Accuracy | F1 Score | ROC-AUC |
|---|---|---|---|
| Random Forest | 0.911 | 0.697 | 0.887 |
| Gradient Boosting | 0.910 | 0.693 | 0.894 |
| XGBoost | 0.909 | 0.694 | 0.885 |
| LightGBM | 0.910 | 0.697 | 0.894 |

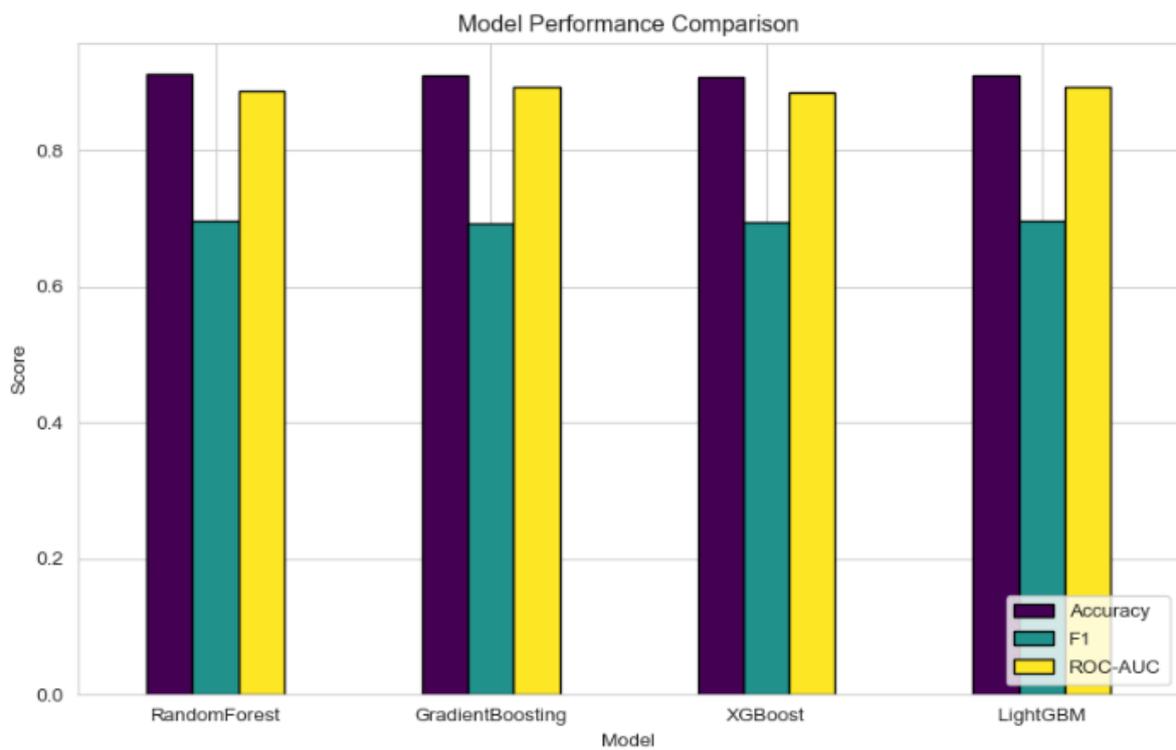

**Fig. 6.** Comparative performance of classification models.

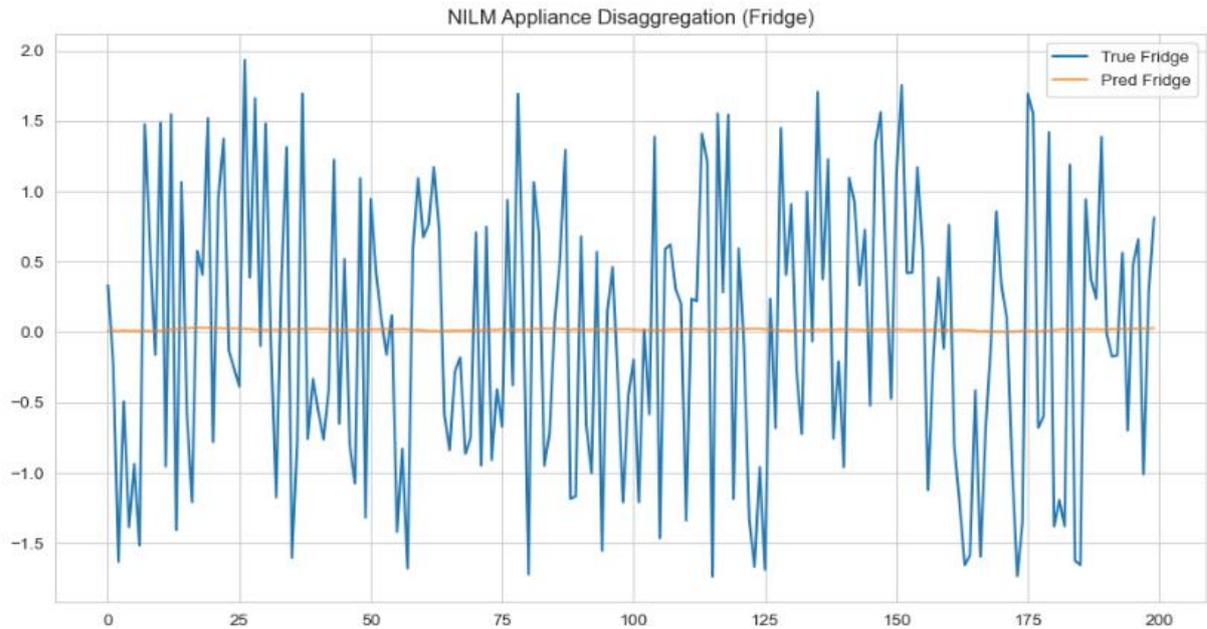

**Fig. 7.** NILM appliance disaggregation result for refrigerator load pattern.

**4.5 Time-Series Anomaly Detection**

The anomaly detection results are illustrated in Fig. 8. The figure shows that anomalies are identified at points where there are abrupt deviations from expected consumption patterns, such as sudden spikes or drops in energy usage. These deviations are indicative of abnormal behavior within the grid. The use of multiple models—namely LSTM, TCN, and Autoencoder—enables the system to effectively capture both short-term irregularities and long-term behavioral changes in the time-series data. By combining these models, the framework leverages their complementary strengths, where LSTM captures sequential dependencies, TCN models long-range temporal patterns, and the Autoencoder detects reconstruction errors associated with anomalies.

The ensemble-based approach further enhances detection performance by reducing false positives. This is achieved by requiring agreement among the models before classifying a data point as anomalous, thereby improving the reliability and robustness of the detection process. To further evaluate the effectiveness of the proposed framework, a pattern recognition analysis was conducted for bypass detection. This stage focuses on identifying abnormal consumption structures that are consistent with meter bypassing, illegal tapping, or concealed energy diversion. Since bypass behavior does not always appear as extreme outliers, it is necessary to compare multiple classification models to determine which algorithm best captures the subtle patterns associated with such activities.

The results of the model comparison, as illustrated in Fig. 9, show that all evaluated classifiers achieved similar accuracy values, indicating that the engineered features provide meaningful discriminatory information. However, a more detailed analysis reveals that ROC-AUC offers better insight into model performance, with Gradient Boosting and LightGBM demonstrating superior ability in distinguishing between normal and suspicious consumption patterns. This suggests that boosting-based models are particularly effective in capturing complex nonlinear relationships associated with bypass behavior.

Despite the strong overall performance, the relatively moderate F1-scores indicate that detecting the minority bypass class remains challenging. This is expected because bypass events are infrequent and often overlap with legitimate variations in consumption, such as low usage periods or irregular household behavior. Consequently, bypass detection is not simply a matter of identifying extreme values, but rather detecting subtle irregularities embedded within normal operational patterns.

The significance of Fig. 9 lies in its ability to visually demonstrate that model selection directly impacts detection reliability. While accuracy values may appear similar across models, the differences become more apparent when evaluated using F1-score and ROC-AUC. This reinforces the importance of using multiple evaluation metrics and justifies the adoption of a hybrid framework, which combines the strengths of different models to achieve more reliable and comprehensive electricity theft detection.

**Table 3.** Performance comparison of models for bypass detection.

| Model | Accuracy | F1 Score | ROC-AUC |
| --- | --- | --- | --- |
| Random Forest | 0.911 | 0.697 | 0.887 |
| Gradient Boosting | 0.910 | 0.693 | 0.894 |
| XGBoost | 0.909 | 0.694 | 0.885 |
| LightGBM | 0.910 | 0.697 | 0.894 |

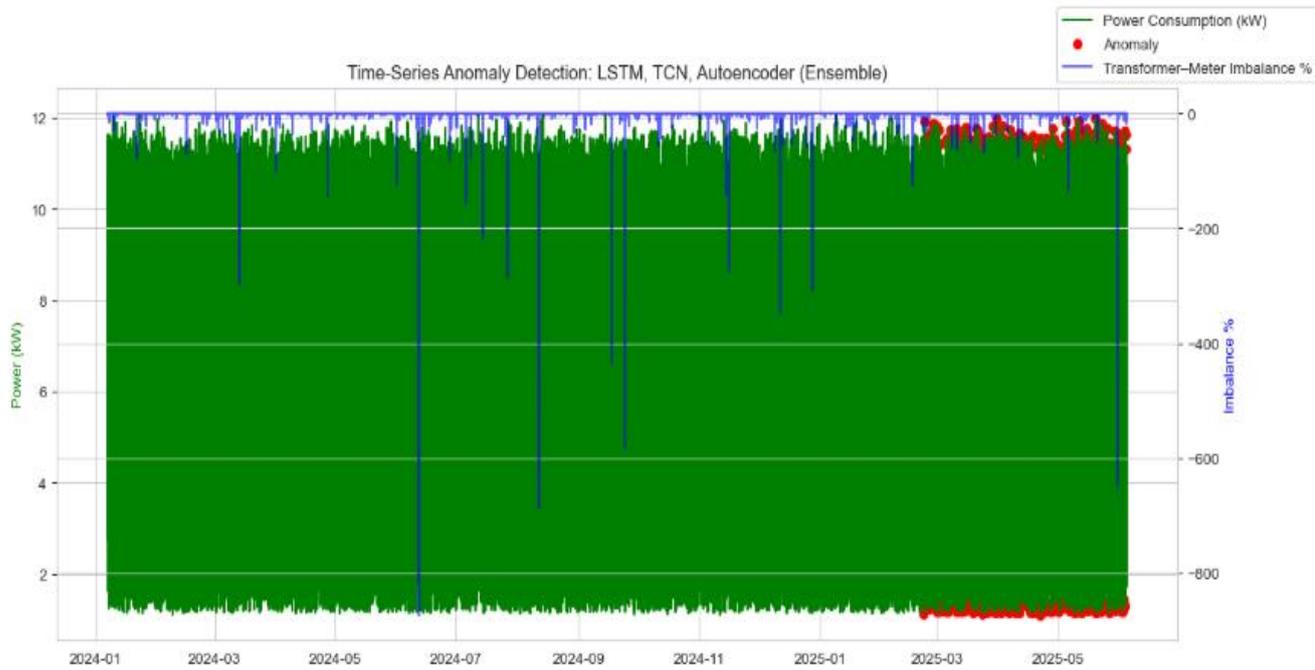

**Fig. 8.** Time-series anomaly detection using deep learning models.

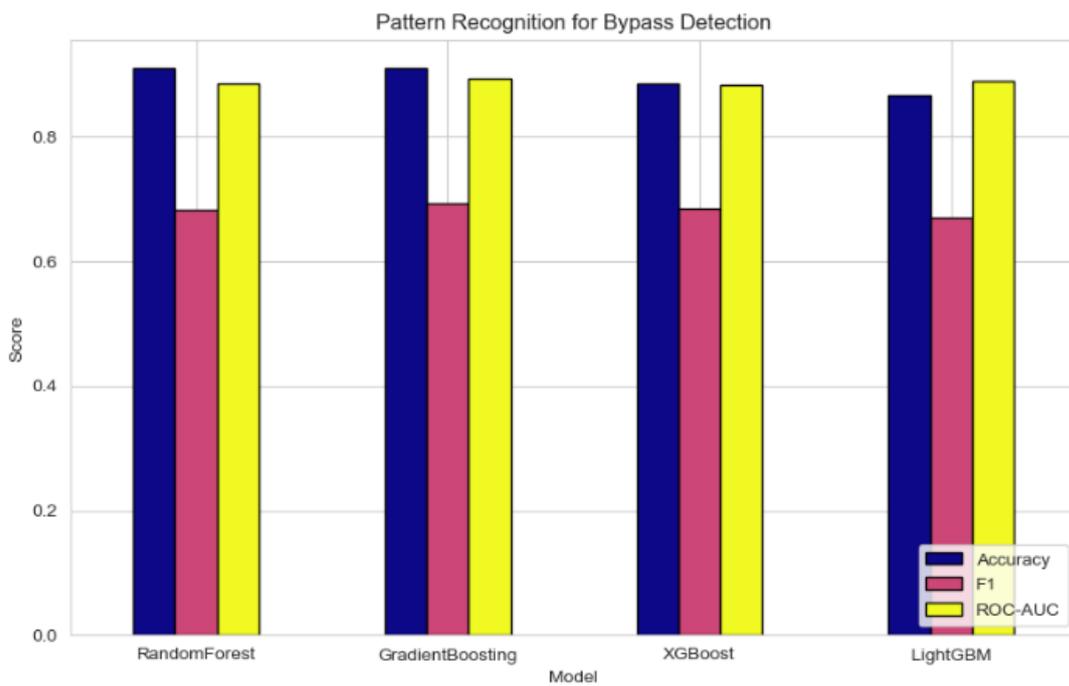

**Fig. 9.** Model comparison for bypass detection pattern recognition.

**4.6 Graph-Based Detection and Network Intelligence**

The results of the graph-based model are presented in Table 4. The GNN model identifies high-risk nodes based on both local features and network relationships. This demonstrates that electricity theft can propagate across connected nodes, making graph-based analysis essential.

The visualization shows how suspicious nodes are distributed within the network, highlighting potential clusters of anomalous activity.

**Table 4. Top suspicious nodes identified by GNN**

| Node | Probability | Label |
| --- | --- | --- |
| Meter_451 | 0.687 | 1 |
| Meter_322 | 0.673 | 1 |
| Meter_81 | 0.671 | 1 |
| Meter_386 | 0.671 | 1 |
| Meter_427 | 0.671 | 1 |

**4.7 Integrated System Performance**

The overall performance of the proposed SGEIS framework is significantly enhanced through the integration of supervised learning, time-series modeling, and graph-based analysis. Rather than relying on a single modeling approach, the system combines multiple complementary techniques, each addressing a specific aspect of the electricity theft detection problem.

Supervised learning models are primarily responsible for classifying known patterns in the data based on labeled examples. These models effectively capture statistical relationships between features and provide strong baseline performance for distinguishing between normal and anomalous consumption. However, their effectiveness is limited when dealing with previously unseen or evolving anomaly patterns.

To address this limitation, time-series models such as LSTM and TCN are incorporated to capture temporal dependencies in electricity consumption. These models are capable of identifying irregular fluctuations, sudden changes, and long-term deviations that may indicate abnormal behavior. As a result, they enhance the system's ability to detect dynamic and time-dependent anomalies that are not easily captured by static classification models.

In addition, graph-based models are employed to represent the structural relationships within the smart grid. By modeling the connections between transformers, feeders, and consumer nodes, these models capture spatial dependencies and enable the detection of coordinated or propagated anomalies across the network. This is particularly important in real-world scenarios where electricity theft may involve multiple interconnected nodes.

The integration of these three components results in a multi-layered detection framework that leverages statistical, temporal, and spatial information simultaneously. This combined approach improves detection accuracy, reduces false positives, and enhances overall system robustness. Compared to standalone models, the hybrid framework provides a more reliable and comprehensive solution for electricity theft detection in complex smart grid environments.

## 4.8 Implications of the Study

The findings of this study provide significant implications for the design and deployment of intelligent electricity theft detection systems in modern smart grids. First, the results demonstrate that relying on a single modeling approach is insufficient for capturing the complexity of electricity consumption behavior. The superior performance of the proposed SGEIS framework highlights the importance of integrating multiple analytical techniques, including machine learning, deep learning, and graph-based models, to achieve robust and reliable detection.

Furthermore, the effectiveness of time-series models such as LSTM and TCN confirms that electricity theft is inherently a temporal problem, where abnormal consumption patterns evolve over time. At the same time, the strong performance of Graph Neural Networks (GNNs) emphasizes that electricity theft is also a spatial phenomenon, often influenced by the structural relationships between meters, transformers, and distribution networks. This dual nature underscores the necessity of combining temporal and spatial modeling for comprehensive grid intelligence.

In addition, the study reveals that feature engineering plays a critical role in improving model performance. Derived features such as supply–demand imbalance, rolling consumption statistics, and environmental adjustments significantly enhance the ability of models to distinguish between normal and anomalous behavior. This suggests that domain knowledge remains essential in designing effective AI-driven smart grid solutions.

From a practical perspective, the proposed SGEIS framework offers a scalable and deployable solution for utility companies. Its ability to detect anomalies in near real-time enables proactive monitoring and early intervention, which can reduce financial losses associated with electricity theft and improve overall grid reliability. Moreover, the integration of graph-based intelligence provides actionable insights into network-level anomalies, enabling utilities to identify high-risk regions and prioritize inspections.

Finally, the incorporation of advanced techniques such as ensemble learning and hybrid modeling contributes to improved system robustness and adaptability. This makes the framework suitable for real-world deployment in dynamic and heterogeneous energy environments, where consumption patterns are influenced by renewable energy integration, environmental conditions, and evolving user behaviors.

Overall, the implications of this study extend beyond electricity theft detection, providing a foundation for the development of next-generation intelligent energy management systems that are efficient, secure, and sustainable.

## 5.0 Conclusion and Recommendations

This study presented the SmartGuard Energy Intelligence System (SGEIS), an integrated artificial intelligence framework designed for electricity theft detection and smart grid intelligence. The proposed system combines supervised machine learning, deep learning-based time-series models, Non-Intrusive Load Monitoring (NILM), and graph-based analysis to address the complex temporal and spatial characteristics of electricity consumption.

The results demonstrate that electricity theft detection is inherently challenging due to class imbalance and the overlap between normal and anomalous consumption patterns. While supervised learning models achieved strong overall accuracy, further evaluation showed that metrics such as F1-score and ROC-AUC provide a more reliable measure of detection performance, particularly for the minority theft class. The integration of time-series models, including LSTM and TCN, enhanced anomaly detection by capturing temporal irregularities and behavioral shifts that are not easily identified by static models.

In addition, the NILM component enabled appliance-level analysis, allowing the system to move beyond aggregate consumption and provide interpretable insights into the sources of abnormal behavior. This is particularly valuable for practical deployment, as it supports utility operators in identifying not only the occurrence of anomalies but also the contributing load patterns. Furthermore, the incorporation of graph-based models allowed the system to capture spatial dependencies within the grid, enabling the detection of coordinated and propagated anomalies across interconnected nodes.

The integration of these components into a unified framework significantly improved detection capability, robustness, and reliability. The findings confirm that a hybrid approach, combining statistical, temporal, and spatial modeling techniques, is essential for effective electricity theft detection in modern smart grid environments.

Based on these findings, several recommendations are proposed. Future research should focus on incorporating real-time data streams to enable continuous monitoring and faster anomaly detection. The use of advanced temporal graph models is also recommended to better capture dynamic interactions within the grid. Additionally, improved techniques for handling class imbalance, such as cost-sensitive learning and advanced resampling strategies, should be explored to enhance detection of rare theft events. Expanding the NILM component to include multiple appliance categories would further improve interpretability and diagnostic capability.

Finally, validation of the proposed framework using real-world utility data is essential to assess its practical effectiveness and scalability.

Overall, the SGEIS framework provides a comprehensive and scalable solution for electricity theft detection, with strong potential for real-world deployment in intelligent energy management systems.

**References**

[1] B. Kabir, U. Qasim, N. Javaid, F. A. Khan, B. Mansoor, A. K. J. Saudagar, H. S. AlSagri, and M. N. Saqib, "Detecting electricity theft in smart grids using optimized machine learning ensemble techniques and eXplainable AI," *Energy Reports*, vol. 11, pp. 1023–1037, 208. [Online]. Available: https://doi.org/10.1016/j.egyr.208.09.019

[2] Iftikhar, N. Khan, M. A. Raza, G. Abbas, M. Khan, M. Aoudia, E. Touti, and A. Emara, "Electricity theft detection in smart grid using machine learning," *Frontiers in Energy Research*, vol. 12, Article 1383090, 2024. [Online]. Available: https://doi.org/10.3389/fenrg.2024.1383090

[3] N. G. Ezeji, K. I. Chibueze, and N. H. Nwobodo-Nzeribe, "Developing and Implementing an Artificial Intelligence (AI)-Driven System for Electricity Theft Detection", *AJERD*, vol. 7, no. 2, pp. 317–328, Sep. 2024.

[4] S. Munawar, N. Javaid, Z. A. Khan, N. I. Chaudhary, M. A. Z. Raja, A. H. Milyani, and A. A. Azhari, "Electricity Theft Detection in Smart Grids Using a Hybrid BiGRU–BiLSTM Model with Feature Engineering-Based Preprocessing," *Sensors*, vol. 22, no. 20, p. 7818, 2022. [Online]. Available: https://doi.org/10.3390/s22207818

[5] I. Petrlik, P. Lezama, C. Rodriguez, R. Inquilla, J. E. Reyna-González, and R. Esparza, "Electricity Theft Detection using Machine Learning," International Journal of Advanced Computer Science and Applications (IJACSA), vol. 13, no. 12, pp. 420–427, 2022. [Online].

Available:https://www.researchgate.net/publication/366750509_Electricity_Theft_Detection_using_Machine_Learning

[6] M. Anwar, S. Abdullah, and R. Ahmed, "Electricity theft detection using pipeline machine learning techniques," in Proc. IEEE Int. Wireless Communications and Mobile Computing Conf. (IWCMC), 2020, pp. 1456–1461.

[7] S. M. Saqib, T. Mazhar, M. Iqbal, A. Almogren, Y. Y. Ghadi, T. Shahzad, and H. Hamam, "Utilizing machine learning ensembles for effective electricity theft detection," *Vol. 44, Issue 1*, Sep. 208. [Online]. Available: https://doi.org/10.1177/0144598781381989

[8] A. I. Kawoosa, D. Prashar, G. R. Anantha Raman, A. Bijalwan, M. A. Haq, M. Aleisa, and A. Alenizi, "Improving Electricity Theft Detection Using Electricity Information Collection System and Customers' Consumption Patterns," *Energy Exploration & Exploitation,* vol. 42, no. 5, pp. 1234–1248, May 2024. [Online]. Available: https://doi.org/10.1177/01445987241255394

[9] X. Chen, C. Huang, Y. Zhang, and H. Wang, "Smart Energy Guardian: A Hybrid Deep Learning Model for Detecting Fraudulent PV Generation," *arXiv preprint arXiv:805.18755 [cs.LG]*, May 208. [Online]. Available: https://doi.org/10.48550/arXiv.805.18755

[10] H. Qu, Y. Lv, X. Luo, Q. Zhu, C. Wu, and J. Pan, "A Privacy-Preserving Federated Learning Approach for Cross-Domain Line Loss Prediction," in *Proc. 208 7th Int. Acad. Exchange Conf. Sci. Technol. Innovation (IAECST)*, pp. 463–467, 208.

[11] C. Collier and K. Guha, "Lightweight LSTM Model for Energy Theft Detection via Input Data Reduction," *arXiv preprint arXiv:807.02872 [cs.LG]*, Jun. 208. [Online]. Available: https://arxiv.org/abs/807.02872.

[12] F. Bibi, S. U. Rehman, S. Bibi, et al., "Reinforcing smart grid resilience through blockchain-supported deep learning models for theft detection," *Scientific Reports*, vol. 16, p. 10515, 2026. [Online]. Available: https://doi.org/10.1038/s41598-026-38824-w

[13] O. Akintola, B. Adetokun, and O. Oghorada, "Robust Energy Theft Detection in Smart Distribution Method Using a Data-driven Method," *Journal of Electrical and Electronic Engineering,* vol. 14, no. 1, pp. 46–53, Feb. 2026. [Online]. Available: https://doi.org/10.11648/j.jeee.20261401.15

[14] Z. Qu, H. Liu, Z. Wang, J. Xu, P. Zhang, and H. Zeng, "A combined genetic optimization with AdaBoost ensemble model for anomaly detection in buildings electricity consumption," *Energy and Buildings*, vol. 248, p. 111193, 2021. [Online]. Available: https://doi.org/10.1016/j.enbuild.2021.111193

[15] A. I. Kawoosa, D. Prashar, M. Faheem, N. Jha, and A. A. Khan, "Using machine learning ensemble method for detection of energy theft in smart meters," *IET Generation, Transmission & Distribution*, vol. 17, no. 21, pp. 4794–4809, Sep. 2023. [Online]. Available: https://doi.org/10.1049/gtd2.12997

[16] M. Irfan, N. Ayub, F. Althobiani, Z. Ali, M. Idrees, S. Ullah, S. Rahman, A. S. Alwadie, S. M. Ghonaim, H. Abdushkour, F. S. Alkahtani, S. Alqhtani, and P. Gas, "Energy Theft Identification Using Adaboost Ensembler in the Smart Grids," *Computers, Materials & Continua*, vol. 72, no. 1, pp. 2141–2158, 2022. [Online]. Available: https://doi.org/10.32604/cmc.2022.08466

[17] I. H. Abdulqadder, I. T. Aziz, and F. M. F. Flaih, "Robust Electricity Theft Detection in Smart Grids Using Machine Learning and Secure Techniques," *International Journal of Advanced Computer Science and Applications*, vol. 13, no. 12, pp. 1021–1032, 2024. [Online]. Available: https://inass.org/wp-content/uploads/2024/10/208022973-2.pdf

[18] M. M. Abou-Elasaad, S. G. Sayed, and M. M. El-Dakroury, "Smart Grid intrusion detection system based on AI techniques," *Journal of Cybersecurity and Information Management (JCIM)*, vol. 15, no. 2, pp. 195–207, 208. [Online]. Available: https://doi.org/10.54216/JCIM.150215

[19] A. S. S. Murugan, R. K. Kar, M. Rajitha, and K. Natarajan, "IoT based Smart Unlawful Electricity Usage Surveillance and Warning System," in *E3S Web of Conferences*, vol. 472, Jan. 2024, Art. no. 03009. [Online]. Available: https://doi.org/10.1051/e3sconf/202447203009

[20] A. Ullah, I. U. Khan, M. Z. Younas, M. Ahmad, and N. Kryvinska, "Robust resampling and stacked learning models for electricity theft detection in smart grid," *Energy Reports*, vol. 13, pp. 770–779, Jun. 208. [Online]. Available: https://doi.org/10.1016/j.egyr.2024.12.041

[21] M. Wen, R. Xie, K. Lu, L. Wang, and K. Zhang, "FedDetect: A Novel Privacy-Preserving Federated Learning Framework for Energy Theft Detection in Smart Grid," *IEEE Internet of Things Journal*, vol. 9, no. 8, pp. 6352–6364, Aug. 2022. [Online]. Available: https://doi.org/10.1109/JIOT.2022.95319

[22] A. U. Oguntola, A. A. Olowookere, M.A Madehin and A. A. Adesope, "Spatio-temporal graph neural network and LSTM hybrid framework for electricity theft detection," arXiv preprint arXiv:2603.20488, 2026.

[23] Y. Himeur, K. Ghanem, A. Alsalemi, F. Bensaali, and A. Amira, "Artificial intelligence based anomaly detection of energy consumption in buildings: A review, current trends and new perspectives," *Applied Energy*, vol. 287, p. 116601, Apr. 2021. [Online]. Available: https://doi.org/10.1016/j.apenergy.2021.116601.

[24] N. M. Elshennawy, D. M. Ibrahim, and A. M. Gab Allah, "An efficient electricity theft detection based on deep learning," *Scientific Reports*, vol. 15, Art. no. 12866, Apr. 208. [Online]. Available: https://doi.org/10.1038/s41598-08-93140-z